\pdfoutput=1

\documentclass[11pt]{article}

\usepackage[]{ACL2023}

\usepackage{times}
\usepackage{latexsym}

\usepackage{tikz}
\usepackage[normalem]{ulem}
\usepackage{latexsym}
\usepackage{epsfig}
\usepackage{graphicx}
\usepackage[linesnumbered,ruled,vlined]{algorithm2e}
\usepackage{xcolor}

\usepackage{multirow}
\usepackage{etoolbox}
\usepackage{balance}
\usepackage{amsmath}
\usepackage{amsfonts}
\usepackage{balance}
\usepackage{pifont}
\usepackage{makecell}
\usepackage{hhline}
\usepackage{booktabs}
\usepackage{booktabs}
\usepackage{makecell}
\usepackage{graphicx}
\usepackage{mathtools}
\usepackage{enumitem}
\usepackage{float}
\usepackage{colortbl}
\usepackage[most]{tcolorbox}
\usepackage{pifont}
\usepackage{subcaption}
\usepackage{hyperref}
\usepackage{algorithmic}
\usepackage{cancel}

\usepackage[T1]{fontenc}

\usepackage[utf8]{inputenc}

\usepackage{microtype}

\usepackage{inconsolata}

\title{Intended Target Identification for Anomia Patients\\ with Gradient-based Selective Augmentation}

\newcommand\correspondingauthor{\thanks{~~Corresponding author.}}
\author{Jongho Kim$^{\spadesuit}$$^{\clubsuit}$, Romain Storaï$^{\spadesuit}$, Seung-won Hwang$^{\spadesuit}$$^{\clubsuit}$\correspondingauthor \\
$^{\spadesuit}$Seoul National University \\
$^{\clubsuit}$ Interdisciplinary Program in Artificial Intelligence, Seoul National University \\ 
\texttt{\{jongh97, romsto, seungwonh\}@snu.ac.kr} \\
}

\newcommand{\ours}{GradSelect\xspace}

\newcommand{\spe}{SPE\xspace}

\newcommand{\jongho}[1]{{\color{black}#1}}

\begin{document}

\maketitle
\begin{abstract}


In this study, we investigate the potential of language models (LMs) in aiding patients experiencing anomia, a difficulty identifying the names of items. Identifying the intended target item from patient's circumlocution involves the two challenges of term failure and error:
(1) The terms relevant to identifying the item remain \textbf{unseen}.
(2) What makes the challenge unique is inherent perturbed terms by \textbf{semantic paraphasia}, which are not exactly related to the target item, hindering the identification process.
To address each, we propose robustifying the model from semantically paraphasic errors and enhancing the model with unseen terms with gradient-based selective augmentation.
Specifically, the gradient value controls augmented data quality amid semantic errors, while the gradient variance guides the inclusion of unseen but relevant terms.
Due to limited domain-specific datasets, we evaluate the model on the Tip-of-the-Tongue dataset as an intermediary task and then apply our findings to real patient data from AphasiaBank. Our results demonstrate strong performance against baselines, aiding anomia patients by addressing the outlined challenges.

\end{abstract}
\section{Introduction}
\label{sec:intro}

Despite significant advancements in language models (LMs), challenges persist in effectively handling the tail data, such as accommodating the needs of unseen language groups and addressing social biases~\cite{gallegos2024bias,10.1162/tacl_a_00615}. This gap underscores the importance of research endeavors focused on refining LMs to better serve underrepresented populations, with individuals having language disorders being no exception.

\begin{figure}[t]
    \centering
    \includegraphics[width=0.98\linewidth]{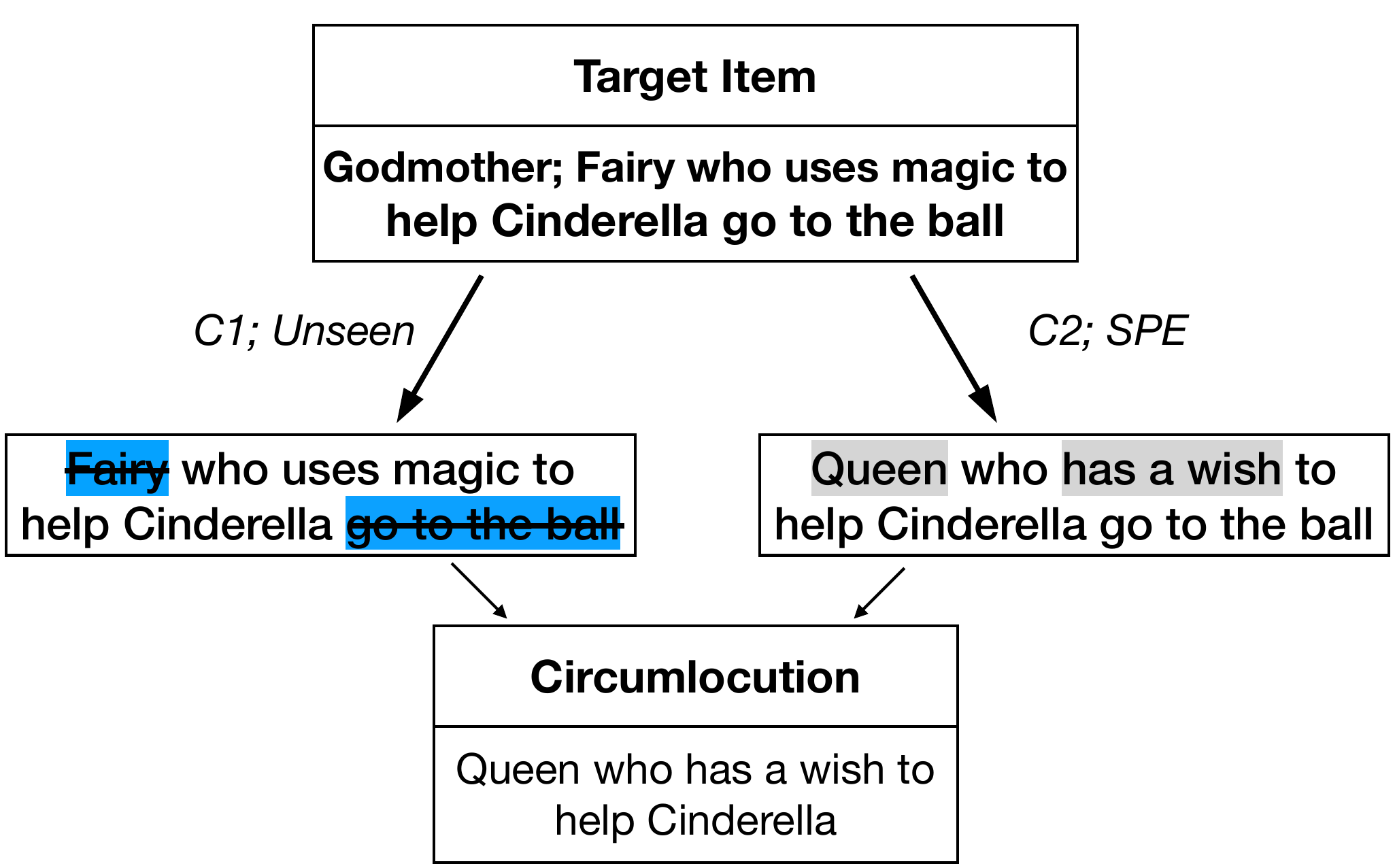}
    \caption{Example of target item identification with circumlocution. Terms with a \sout{\textcolor{blue}{blue}} background are unseen, and those with a \textcolor{gray}{gray} background are SPE.}
    \label{fig:examplesdataset}
    \vspace{-5mm}
\end{figure}

Anomia or word-retrieval difficulty stands as one of the most prevalent symptoms of People With Aphasia (PWA)~\cite{laine2013anomia}. 

Anomic individuals typically experience tip-of-the-tongue phenomenon~\cite{goodglass1976tip}, where they are aware of the target item they want to convey but face difficulty in retrieving suitable words to articulate it.
\jongho{This difficulty frequently appears as `circumlocution'~\footnote{\url{https://aphasia.org/what-is-aphasia/}}, where individuals talk around the word.
They rely on terms with paraphasic errors, or word substitutions, producing maybe-related or completely unrelated words~\cite{friedman2015encyclopedia}}.

In this study, we aim to design an LM that assists anomia patients by identifying their intended target item: For the given circumlocution of the individual, the model should identify the target item from the corpus. 
Surprisingly, while anomia significantly impacts the ability of individuals to engage in meaningful conversations~\cite{code1999emotional}, there is no such LM specifically designed for assistance.
Primitive works focused only on evaluating the LMs' performance of intended target identification~\cite{purohit2023chatgpt, salem2023automating}. Moreover, current LMs fail to suggest intended items, as will be shown in Sec.~\ref{sec:experiments}, further highlighting the need for improvements in this area.

We start by specifying the two challenges from the anomic speech. 
\begin{itemize}
    \item C1-Word retrieval failure: Individuals fail to recall the relevant terms and can only provide limited information about the target item, so the relevant terms are \textbf{unseen} in the circumlocution~\cite{puttanna2021treatment}. 
    \item C2-Word retrieval error: Individuals make errors in word usage. As anomia is linked with a disorder of losing semantic knowledge about object concepts, it leads to the production of perturbed terms with \textbf{semantically paraphasic errors (\spe)} when attempting to name those concepts~\cite{reilly2011anomia, harnish2018anomia, salem2023refining, binder2009encyclopedia}.
\end{itemize} 

C1 is a challenge that is commonly faced in search, and thus relatively well-studied, aligning with the works revealing LMs' vulnerability to incomplete inputs~\cite{yu2021improving, wang2023colbert, mackie2023generative}. \jongho{The example is shown in the left part of Fig.~\ref{fig:examplesdataset}. The relevant description, such as \textit{`fairy'} or \textit{`go to the ball'}, is required to identify the target item \textit{`Godmother'}, but these terms are unseen in the circumlocution.}

A more unique challenge to anomia is C2: its inherent perturbation from SPE,
which may cause the model to identify the wrong item.
For example in the right part of Fig.~\ref{fig:examplesdataset}, the individual uses words such as \textit{queen}, which are perturbed about the target item.
Such \spe terms are not semantically related to the target item and therefore do not assist in the model's identification process; they may even be detrimental. 
Specifically, our pilot study found that roughly 40\% of the terms in the circumlocution degrade the model performance. 
Therefore, anomia presents a complex challenge where we must navigate the unseen terms (C1) amidst the innate presence of \spe (C2).

\begin{figure*}[t]
    \centering
    \scalebox{0.88}{
    \includegraphics[width=1.0\linewidth]{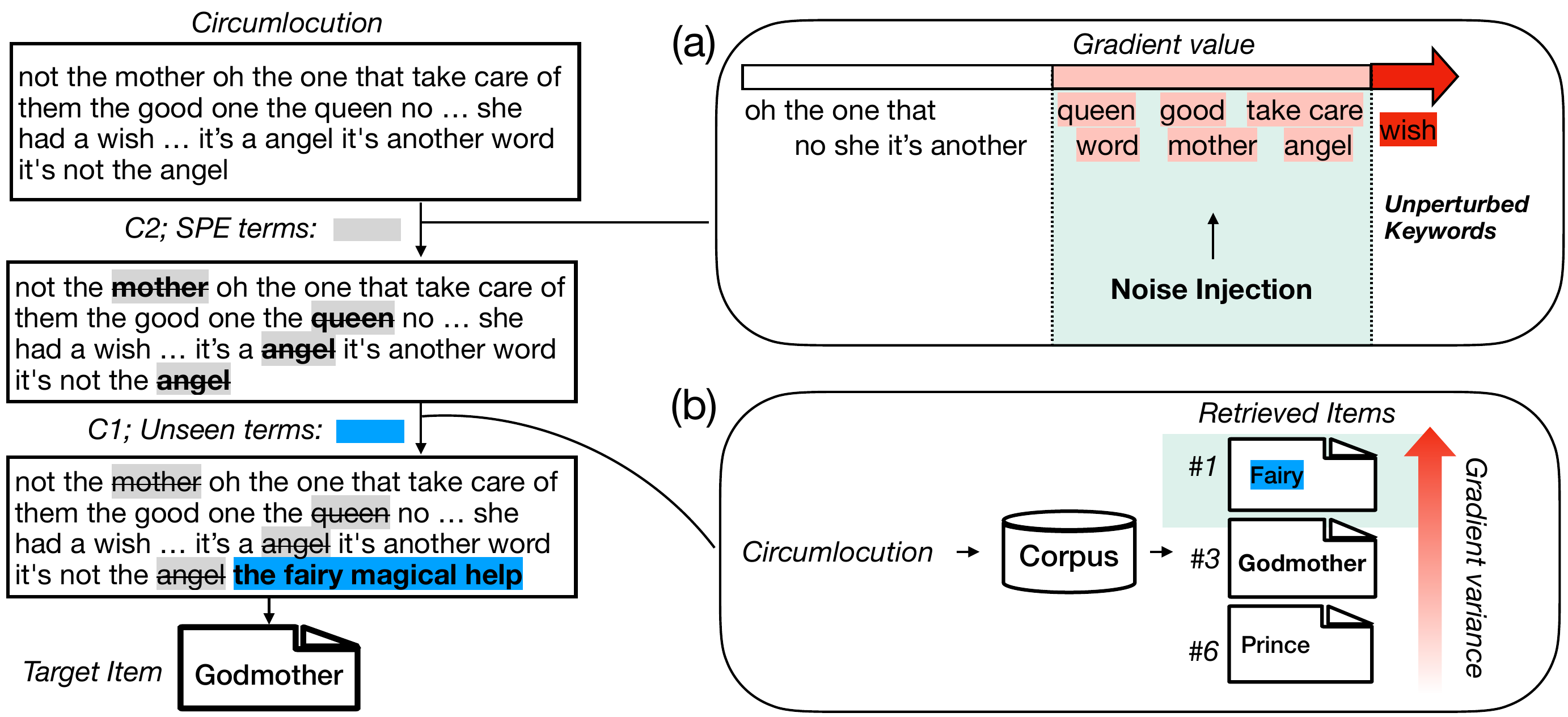}
    }
    \caption{We augment the dataset by leveraging the gradient to select terms to augment. Left side: the goal of \ours. We delete the \textcolor{gray}{SPE terms} while expanding the \textcolor{blue}{unseen terms}. Right side: the description of the selection process. (a) We robustify the model with the noise-injected circumlocution guided by the gradient value. (b) We enhance the representation of the circumlocution with the relevant items based on the gradient variance.}
    \label{fig:solution}
    \vspace{-5mm}
\end{figure*}

To this end, we introduce a novel augmentation approach involving gradient-based selection of augmentation target, called \ours.
The goal is described in color on the left side in Fig.~\ref{fig:solution}. We will delete \textcolor{gray}{the SPE terms} while expanding the \textcolor{blue}{unseen terms}. 

To delete \textcolor{gray}{the \spe terms} (C2), we take an adversarial approach: By injecting more noise into the circumlocution, we robustify the model against \textit{diverse} \spe terms.
However, the challenge lies in that the inherently perturbed circumlocution easily loses its \textit{relevance} to its original target after noise injection.
Our contribution is to control the quality of data that ensures both \textit{diversity} and \textit{relevance}~\cite{ash2019deep}, by assessing the gradient value of each term to select the target for injecting noise. The process is described in the Fig.~\ref{fig:solution}-(a).  
While we inject noise into \textcolor{red!40!white}{high-gradient} terms important to diversify the model's representation, we prevent noise from affecting \textcolor{red}{top-n gradient} terms. This is based on our core finding that such terms are usually unperturbed keywords crucial for maintaining relevance to the correct item.

From the denoised circumlocution, we then address the \textcolor{blue}{relevant but unseen} terms (C1) 
by taking inspiration from pseudo-relevance feedback (PRF)~\cite{croft2010search,lavrenko2017relevance}.
The process follows Fig.~\ref{fig:solution}-(b).
To expand unseen terms (e.g. \textit{fairy}) to seen terms, we augment the target items using the top retrieved items from the initial prediction. Here, we select the candidate items ranked higher than the target item. It stems from the observation that items with relevant terms exhibit a high gradient variances, which can be approximated by their relative rank~\cite{zhou2022simans}.

Our exploration of this methodology begins with the Tip-of-the-Tongue dataset~\cite{bhargav2022s,arguello2023overview}, due to the scarcity of real-world datasets that precisely target anomia. 
Subsequently, we apply and validate our findings using real patient data 
from A-cinderella~\cite{salem2023automating}, encompassing both the original dataset and our custom challenge set.
The results demonstrate that \ours can improve identification accuracy by effectively controlling the quality of augmented data.
\section{Pilot Study on Circumlocution Terms}

This section discusses the existence and effect of each C1: relevant-unseen and C2: seen-\spe term in the circumlocution.

\subsection{Data Source and Models}
It is difficult to directly evaluate the effectiveness of our method on real-world anomic patients due to the scarcity of such datasets tailored to our specific task. \jongho{Therefore, we conduct a pilot study on the TREC-TOT 2023 movie retrieval task~\cite{arguello2023overview}, which involves identifying target movie based on circumlocutions from individuals experiencing the `Tip of the Tongue' phenomenon, a temporary form of anomia. The query often contains incomplete and dummy information from false memories, similar to our anomia scenario.

We used lexical retriever BM25~\cite{robertson2009probabilistic}, and dense retriever co-Condenser~\cite{gao-callan-2022-unsupervised} for the pilot study. BM25 is a traditional method of information retrieval (IR) that relies on exact term matching to find the target document from the query. On the other hand, dense retriever uses dense vector representations for queries and documents and relies on capturing semantic similarity rather than exact term matches. Co-Condenser is one variant of dense retriever, which additionally pre-trains dense retriever with corpus-level contrastive loss. We use the co-Condenser version from~\citet{Kim2023OnIT}~\footnote{\citet{Kim2023OnIT} pre-trained the dense retriever on the Wikipedia corpus under the audiovisual works domain and used MaxSim operator during the fine-tuning stage.} and refer to it as `co-Condenser*'.

The normalized discounted cumulative gain (nDCG) score~\cite{jarvelin2002cumulated} is used to evaluate the performance of IR models.
}

\subsection{Relevant but Unseen Terms}

\begin{table}[ht]
\centering
\begin{tabular}{l|c}
\toprule
\textbf{Dataset} & nDCG@10 \\ \midrule
MS Marco & 0.228 \\ 
BEIR (average) &  0.423 \\ \midrule
TREC-TOT &  0.093 \\ \bottomrule

\end{tabular}

\caption{The BM25 performances that confirm the challenge of unseen terms. nDCG@10 score is reported.}
\label{tab:tomt_bm25}
\end{table}


\jongho{The definition and the presence of unseen terms are determined through the simple rule: `Seen terms' refer to the terms that appear in the circumlocution, while `unseen terms' are those that do not appear in the circumlocution.}

The previous work found the lexical overlap between the circumlocution and the target item is lower in Tip-of-the-Tongue movie domain compared to conventional IR benchmarks (e.g. MSMarco~\cite{nguyen2016ms}: 0.55 vs Tip-of-the-Tongue movie~\cite{bhargav2022s}: 0.25), and the same trend is also reported in the domain of book and music~\cite{Bhargav2023WhenTM,lin2023decomposing}. 

Furthermore, we compare the performances of BM25 on TREC-TOT with other datasets. Other datasets include MSMarco~\cite{nguyen2016ms}, and BEIR~\cite{Thakur2021BEIRAH}, which is the collection of 18 IR datasets.
The results are reported in Table~\ref{tab:tomt_bm25}. The performance of BM25 on TREC-TOT is far behind the other dataset, which indicates the challenge of unseen terms.

\subsection{Seen but \spe Terms}

\begin{table}[ht]
\centering
\scalebox{1.0}{
\begin{tabular}{l|cc}
\toprule
\textbf{Performance Change} & \textbf{Ratio} \\
\midrule
Improved (↑) & 40.1\% \\
Decreased (↓) & 59.9\% \\
\bottomrule
\end{tabular}
}
\caption{Impact of random sentence deletion on model performance. We measure the change of nDCG on the test set of the TREC-TOT.}
\label{tab:tomt_perturb}
\end{table}

\jongho{We define `\spe terms' as those that are not semantically related to the target item and explore how these terms undermine the process of target identification. In this paper, classifying which terms in the circumlocution are considered \spe is model-dependent. If a term is semantically related to the target item, it aids the model in correctly identifying the target item. Conversely, if a term has \spe, it either does not assist the model or could even hinder its performance. The existence of these perturbed terms is confirmed through both quantitative and qualitative methods.}

For the quantitative analysis, we evaluated the change in the performance of the semantic retriever, co-Condenser*, by deleting part of the circumlocution. Our key point is that if the circumlocution suffers from the \spe terms that negatively impact the retrieval procedure, there will be instances where deleting these terms improves model performance.

\jongho{Specifically, we start by filtering the completely unrelated sentences in the circumlocution. TREC-TOT dataset provides sentence-level annotations indicating whether a sentence is about the movie or not. The latter type of sentence includes social words (e.g., \textit{Thanks}) or details about the context in which the movie was watched (e.g., \textit{with my 6-year-old nephew}). With these annotations, we filtered 18.7\% of the sentences that are completely unrelated to the target movie.
Then we measured how deleting each sentence in the circumlocution affects the performance of co-Condenser* on the filtered TREC-TOT test set. The results in Table~\ref{tab:tomt_perturb} imply that 40.1\% of the sentences in the dataset include \spe terms, causing the model to predict the wrong item.}


We further qualitatively confirm that the \spe terms hallucinate the model to identify the wrong item.
We selected some queries for which the model could not identify the correct target item and manually deleted some terms that we considered perturbed. 
The case study example is shown in Appx.~\ref{sec:case_study_perturbed}.
By doing so, we enabled the retriever to rank the target item as the top-1, whereas with the \spe terms it was ranked lower than top-10. It implies that the \spe terms significantly drop the model performance, which implies that robustifying the model from such terms is necessary.
\section{Methods}

Our goal is to identify the intended target item from the item corpus given the circumlocution. 
The flow of \ours is depicted in~Alg.~\ref{alg}. Leveraging the gradient as a proxy for the \spe, \ours selectively augments the dataset to make the semantics between circumlocution and the target item properly overlap.  It is designed to denoise the \spe terms in the seen terms (Subsec~\ref{subsec:seenunintend}), and enhance the circumlocution with unseen but relevant terms in our task (Subsec~\ref{subsec:intendunseen}). 
\begin{algorithm}[h]
\small
 \caption{\ours}
 \label{alg}
 \begin{algorithmic}[1]
 \jongho{
\REQUIRE Circumlocution $C$, item $I$, training dataset $\mathcal{T}=\{(C,I_{+})\}$, teacher model $\Theta_{t}$, student model $\Theta_{s}$
\ENSURE Prediction for the intended target item

\# Circumlocution Augmentation
\STATE Initialize: $\Theta_{t}$ with parameters $\theta$
\FOR{$C \in \mathcal{T}$} 
    \STATE $C = \{c_1, c_2, ..., c_i, ..., c_l\}$ \hfill \# Token list of $C$
    \STATE Compute the importance score $\text{IMP}_{c_i}$ for each $c_i$ using gradient values \hfill \# Refer to Eq.~(\ref{eq:importance_score})
    \STATE Rank the tokens in $C$ in descending order of importance and select a subset $C[m:n]$
    \STATE Apply augmentation targeting $C[m:n]$ to generate $C_{aug}$
    \STATE Calculate the loss and update $\Theta_{t}$ \hfill \# Refer to Eq.~(\ref{eq:cutoffloss})
\ENDFOR

\# Item Augmentation
\STATE Initialize: new training set $\mathcal{T'}$
\FOR{$C \in \mathcal{T}$}
    \STATE Get the ranked list $R$ by predicting the target item for $C$ with $\Theta_{t}$
    \STATE $r_g \leftarrow$ Rank of the intended target item in $R$
    \IF {$r_g > k$}
        \STATE Add top-$k$ items $(C,I'_{+})$ from $R$ to $\mathcal{T'}$
    \ENDIF
\ENDFOR

\# Student Model Training
\STATE Initialize: $\Theta_{s}$ with parameters $\theta$
\FOR{$C \in \mathcal{T} \cup \mathcal{T'}$}
    \STATE Repeat the subprocedure of Circumlocution Augmentation (lines 3-7) using $\Theta_{s}$
\ENDFOR

\# Model Evaluation
\STATE Get predictions with $\Theta_{s}$ on the test set
\RETURN Ensembled predictions from $\Theta_{s}$ and $\Theta_{t}$
}
\end{algorithmic}
\end{algorithm}

\subsection{Deleting Seen but \spe Terms}
\label{subsec:seenunintend}

We first target the seen but \spe terms by the adversarial approach:
We expose the model to diverse forms of \spe terms, forcing it to learn terms that are crucial for accurate prediction, rather than relying on \spe terms.
The challenge is that as the circumlocution is inherently perturbed by \spe, unconstrained noise injection might leave only \spe terms with no relevance to the target item.
To overcome this, we propose to control the quality of the data by selecting the pool of the terms to be noised. \jongho{We focus on augmenting data that are semantically \textit{diverse}-covering a wide range of expressions-but still \textit{relevant}-remaining pertinent to the target-which are the two measures of data quality~\cite{ash2019deep,zhao-etal-2022-epida}.}

\jongho{\subsubsection{Circumlocution Augmentation}}
Our contribution is that we leverage the gradient value of each term to select the augmentation target, balancing diversity and relevance.
In essence, both the \spe and relevant terms will significantly impact the model's prediction of the item. The high gradient value of a term indicates such an impact. Our key finding is that during the train time, when the model is reliable, top gradient terms are the unperturbed keywords that are essential for predicting the accurate item, while \spe terms are likely to have a less pronounced impact on accurate prediction (Subsec.~\ref{subsec:gradientasaproxy}). Therefore, while we select to noise the terms that affect the model performance for diversity, we leave keywords untouched to preserve the semantic relevance.

The selection algorithm is explained in Alg.~\ref{alg} lines 1-8.
Let $C$ be the input circumlocution with $[c_i]^l_{i=1}$ tokens and $C^h = [c^h_i]^l_{i=1}$ the input embedding matrix.
We augment the train data from $C$ to $C_\text{aug}$ by noising vectors along the token axis ($l$).

We first compute the importance of each embedding by calculating the gradient magnitude following~\cite{li2019textbugger, wu2023prada}. We sum up the scores across the hidden dimension in the embedding space to obtain a token-level importance score. The scoring function that assesses the importance of $i$-th token, $\text{IMP}_{c_i}$, is
\begin{equation}
\label{eq:importance_score}
    \text{IMP}_{c_i} = \left\|\frac{\partial \mathcal{F}_{I}(C)}{\partial \boldsymbol{c}^{h}_{i}}\right\|^2_2
\end{equation}
where $\mathcal{F}_{I}(C)$ represents the model's prediction of relevance scores for the items.
We rank all tokens based on their importance score $\text{IMP}_{c_i}$ in descending order.
The bottom $n$ tokens with the low gradients are mostly stop words~\cite{wang2020gradient} or don't affect the semantics. By focusing on noising high-gradient terms ($C[:n]$), we effectively create diverse meanings of circumlocution. Then, we focus on preserving the top $m$-terms for each circumlocution so that the key relevant terms are retained.
Therefore, the resulting noise exclusively targets the tokens within the $C[m:n]$ range. 



Then, we augment the selected tokens in the circumlocution by injecting more noise. This noise can be introduced by either adding random noise to the token embedding~\cite{zhou-etal-2021-virtual}, or by deleting part of the embedding~\cite{shen2020simple}.

The loss function utilized for training is defined following Cutoff~\cite{shen2020simple}:
\begin{align}
\label{eq:cutoffloss}
\mathcal{L}= & \ {\mathcal{L}_{\rm ce}(C, I)} +  \alpha{\mathcal{L}_{\rm ce}(C_{\text{aug}}, I)} \nonumber \\
& +\ \beta\mathcal{L}_{\rm js}(C, C_\text{aug}) 
\end{align}
where $\mathcal{L}_{\rm ce}$ represents the cross-entropy loss and $\mathcal{L}_{\rm divergence}$ the Jensen-Shannon (JS) divergence consistency loss. 


\subsection{Expanding Relevant but Unseen Terms}
\label{subsec:intendunseen}
\jongho{
While promising, there remains room for improvement in the relevant but unseen dimension of the circumlocution. To this end, we propose to augment the target items that may contain the unseen-relevant terms from top-ranked candidates.
The idea aligns with pseudo-relevance feedback (PRF)~\cite{croft2010search,lavrenko2017relevance}, which aggregates top retrieved items from an initial search to original query embedding to capture the unseen information for better query representation.} Our distinction is that we address the risk of naive PRF that expands irrelevant items together~\cite{li2022does}, by selectively distilling the relevant item with gradient variance.



\subsubsection{Target Item Augmentation}

We propose to leverage the variance of gradients to better distill the items with relevant terms. If one item has the relevant terms, it will be semantically similar to the target item but annotated as negatives, which exhibit high gradient variance~\cite{agarwal2022estimating,zhou2022simans}.

\jongho{
Following SimANS~\cite{zhou2022simans} that used the relative relevance scores to substitute the time-consuming gradient variance computation, 
our idea is to selectively extract the potentially relevant items based on the relative rank concerning the original target item as described in Alg.~\ref{alg} lines 9-16. We denote $C$ as the circumlocution and $I_{+}$ as the corresponding target item. With the model $\Theta_{t}$ trained on the original dataset $\mathcal{T}={(C,I_{+})}$, we first choose circumlocutions for which the target item is not ranked in the top-$k$ retrieved items. We regard such circumlocutions as those requiring unseen terms. Then, we extract items $I'_{+}$ whose rank is higher than $k$ as additional target items that provide relevant terms. As a result, we get the new training set $\mathcal{T'}=\{(C,I'_{+})\}$.
 
To leverage the new dataset, we distill the knowledge of items with relevant terms by self-knowledge distillation (KD)~\citet{furlanello2018born}. Self-KD is a technique where a neural network improves its performance by using its own outputs as training labels, serving as both the teacher and student models. 
Our procedure is explained in Alg.~\ref{alg} lines 17-22.
The student model is newly initialized with the same parameters before the teacher is trained on $\mathcal{T}$, and the training process outlined in SubSec.~\ref{subsec:gradientasaproxy} is repeated using the augmented dataset $\mathcal{T} \cup \mathcal{T'}$. By learning from such a dataset, the student model can understand the relevance of target items that were originally unseen by the teacher model. The final prediction is the \jongho{ensembled} prediction of $\Theta_{t}$ and $\Theta_{s}$ following~\citet{furlanello2018born}.}

 

\section{Experiments}
\label{sec:experiments}

\begin{table*}[t]
\centering
\scalebox{0.9}{
\begin{tabular}{l|cccc|cccc}
\toprule
\multicolumn{1}{c|}{\multirow{2}{*}{\textbf{Models}}} & \multicolumn{4}{c|}{\textbf{Reddit-TOMT}} & \multicolumn{4}{c}{\textbf{TREC-TOT}} \\
& nDCG & nDCG@10 & R@1 & MRR & nDCG & nDCG@10 & R@1 & MRR \\ \midrule
DPR & 0.5515 & 0.4955 & 0.3564 & 0.4557 & 0.1797 & 0.0954 & 0.0533 & 0.0873 \\
co-Condenser*  & 0.5997 & 0.5526 & 0.413 & 0.5121 & 0.3109 & 0.2341 & 0.1467 & 0.2099 \\ \midrule
\multicolumn{9}{c}{+ Circumlocution augmentation} \\ \midrule
Cutoff & 0.6064 & 0.5579 & 0.425 & 0.5202 & 0.3042 & 0.2306 & 0.12 & 0.1996 \\
$\text{\ours}^{d}_{C}$ & \textbf{0.6135} & \textbf{0.5673} & \textbf{0.4295} & \textbf{0.528} & \textbf{0.3377} & \textbf{0.2695} & \textbf{0.1667} & \textbf{0.2408} \\ \midrule
\multicolumn{9}{c}{+ Target item augmentation} \\ \midrule
hard-label KD & 0.6284 & 0.5808 & 0.4446 & 0.5416 & 0.3299 & 0.2543 & 0.16 & 0.2322 \\
soft-label KD & 0.6233 & 0.5778 & 0.4454 & 0.5403 & 0.3405 & 0.2677 & 0.1667 & 0.2445 \\
$\text{\ours}^{d}$ & \textbf{0.6296} & \textbf{0.5868} & \textbf{0.4514} & \textbf{0.5478} & \textbf{0.3413} & \textbf{0.2698} & \textbf{0.1733} & \textbf{0.2456} \\ 
\bottomrule
\end{tabular}
}
\caption{nDCG@1000, nDCG@10, Recall, and MRR on both Reddit-TOMT and TREC-TOT test sets. We incrementally add our component and compare it with the baselines. The best scores are highlighted in bold.}
\label{tab:main_tot}
\vspace{-4mm}
\end{table*}

 In this section, we first evaluate our strategy for the intermediary task that simulates item recall difficulties. Subsequently, we transition to use the utterances from real-world PWA datasets sourced from AphasiaBank~\cite{forbes2012aphasiabank}, which allows us to validate our findings in a more clinically relevant context.

\subsection{Known-item Retrieval}
\label{subsec:known-item-IR}

\paragraph{Dataset and Evaluation Details}
Reddit-TOMT~\cite{bhargav2022s} and TREC-TOT 2023~\cite{arguello2023overview} are information retrieval benchmarks involving the retrieval of a target movie for which a user cannot recall a precise identifier. Compared to Reddit-TOMT, TREC-TOT 2023~\cite{arguello2023overview} consists of smaller queries and a huge corpus pool. We leverage it to verify the effectiveness of our approach with varying data sizes. Details on data statistics are in Appx.~\ref{appndx:eval_details}.

For evaluation, we build our backbone model \textbf{co-Condenser*} from~\citet{Kim2023OnIT}, where co-Condenser~\cite{gao-callan-2021-condenser} which is pre-trained in domain-specific corpus and the MaxSim operator handles documents exceeding the model's token limit. 
We inject the noise by random deletion (\textbf{$\text{\ours}^{d}_{C}$}) and incrementally apply selective Self-KD (\textbf{$\text{\ours}^{d}$}).
We evaluated the test sets using three standard metrics in the retrieval, namely nDCG, Recall (R), and the Mean Reciprocal Rank (MRR).
More details on data and settings are in Appx.~\ref{appndx:eval_details}.

\paragraph{Baselines} 

We compare \ours with the approaches that our method builds upon. We use
(1) Circumlocution augmentation: \textbf{Cutoff}~\cite{shen2020simple} for original random deletion performance.
Upon $\text{\ours}^{d}_{C}$, we implement (2) Item augmentation with Self-KD~\cite{furlanello2018born} methods with both \textbf{soft} and \textbf{hard} labels to highlight the benefits of our enhancement.

\paragraph{Results}

The results on Reddit-TOMT and TREC-TOT 2023 are presented in Table \ref{tab:main_tot}. \ours improves performance on both datasets, with both components of \ours proving effective.
For the circumlocution augmentation, $\text{\ours}^{d}_{C}$ outperforms Cutoff, achieving higher scores across all metrics. 
Moreover, the incremental application of our item augmentation further boosts \ours. Our strategy of selecting relevant items consistently achieves better performance compared to self-KD with soft and hard labels.

\subsection{Word completion for PWA}

\paragraph{Dataset and Evaluation Details}
A-Cinderella~\cite{salem2023automating}, derived from the transcripts of AphasiaBank~\cite{forbes2012aphasiabank},  is a dataset designed for predicting intended words in cases of paraphasias., The dataset consists of utterances from patients with aphasia (PWA) recalling the story of Cinderella. 
\textit{Paraphasia} is a broader symptom of anomia and refers to \textit{the production of unintended or incorrect words}. Within this dataset, instances of paraphasia are addressed through the procedure: Upon an unintended word by a patient, the word is masked, followed by the masked word prediction to identify the intended word. 

We hypothesize that assessing our approach on this dataset will demonstrate its practical applicability. 
While \textit{anomia} relates specifically to \textit{difficulties in word recall}, the task at hand can also be viewed as relevant to anomias. In this context, the masked word serves as the target word that an anomic patient struggles to recall, with the model assisting in identifying the word.

Additionally, we introduced an additional challenge set that enhances the dataset's relevance to anomia. This is done by removing the ``retracing'' words near the masked item, which could potentially leak the answer during circumlocution~\footnote{About 26\% of the intended targets for paraphasia are retracing, where a speaker reiterates the segment of speech~\cite{salem2023automating} (\textit{e.g. Cinderella tried on the <masked item> slipper slipper}).}.
To simulate the scenario of anomia, the word must not be explicitly included in the circumlocution. Consequently, we delete any instance of the intended word from the context surrounding the masked item to build the challenge set.

The evaluation is done with a 10-fold cross-validation setting. we use \textbf{DPR} as our backbone model. Following the setting of Subsec.~\ref{subsec:known-item-IR}, we also applied the MaxSim operator (\textbf{DPR$_{maxsim}$}) and implemented ours on top of that. We evaluated both version of inserting noise by replacement (\textbf{$\text{\ours}^{r}$}) and deletion (\textbf{$\text{\ours}^{d}$}).
The metrics are exact match (EM) and accuracy at 5 (acc@5) following~\citet{salem2023automating}.

\paragraph{Baselines}

In addition to baselines in Subsec.~\ref{subsec:known-item-IR}, we consider strong augmentation baselines. We compare two types of circumlocution augmentation, replacement and deletion. \textbf{EDA}~\cite{wei2019eda} introduce random insertions, replacements, and deletions together. 
For replacement \textbf{(R)}, we use \textbf{Random} for random noise insertion, \textbf{SMART}~\cite{jiang2020smart} for virtual adversarial noise insertion to create diverse data, \textbf{VDA}~\cite{zhou-etal-2021-virtual} for virtual augmentation strategy that targets both semantic relevance and diversity.
For deletion \textbf{(D)}, in addition to \textbf{Random (Cutoff)}, we use \textbf{Large-loss} that selects the augmented data with a higher loss than the original loss for diversity, as suggested by \citet{Yi2021ReweightingAS,kamalloo-etal-2022-chosen}~\footnote{${\mathcal{L}_{\rm ce}(C, I)} < {\mathcal{L}_{\rm ce}(C_{\text{aug}}, I)}$ in Eq.~\ref{eq:cutoffloss}}.
Conversely, \textbf{Small-loss} selects the data with the lower loss to exclude low-relevance samples~\cite{han2018co}.
Additionally, we include the performance of \textbf{GPT-4}~\cite{openai2023gpt4} for a comprehensive comparison. \jongho{The details of baseline implementations are in Appendix.~\ref{appndx:baseline_details} and~\ref{appndx:gpt4_details}.}

\paragraph{Results}

\begin{table}[t]
\centering
\scalebox{0.75}{
\begin{tabular}{ll|cc|cc}
\toprule 
            &                   & \multicolumn{2}{c|}{\textbf{Original set}} & \multicolumn{2}{c}{\textbf{Challenge set}} \\ \midrule
        & Models                 & EM                  & Acc@5        & EM                  & Acc@5        \\ \midrule \midrule
        & DPR               & 0.3994               & 0.6117     & 0.2860        & 0.4963     \\
        & DPR$_{maxsim}$        &    0.4043       &     0.6094       & 0.3030        & 0.5027     \\ \midrule \midrule
& \multicolumn{5}{c}{+ Circumlocution augmentation} \\ \hline
        &  EDA             &    0.4084         &   0.6099     &     0.2869      &    0.4927   \\ \midrule
\multirow{4}{*}{R}  & Random       &     0.4039       & 0.6150       &       0.2935   &     0.5027   \\
                    & SMART           &    0.4069         &   0.6134     &     0.2918      &    0.5038   \\
                    & VDA           &    0.4072         &   0.6144     &     0.2964      &    0.5052   \\
                    &$\text{\ours}_{C}^{r}$ &   \textbf{0.4238}   & 0.6241       &   0.3203      & 0.5263 \\ \midrule
\multirow{4}{*}{D}  & Random (Cutoff)        & 0.4180            & \textbf{0.6255}       &   0.3237        & 0.5201     \\
                    & Large-loss          & 0.4185            & 0.6191       &   0.3121        & 0.5177     \\
                    & Small-loss          & 0.4161            & 0.6170       &   0.3115        & 0.5132     \\
                    & $\text{\ours}_{C}^{d}$ &      0.4219   &     0.6218       &  \textbf{0.3256}      & \textbf{0.5275} \\ \midrule \midrule
& \multicolumn{5}{c}{+ Target item augmentation} \\ \hline
                    & hard-label KD   &       0.4248    &    0.6197      &         0.3244           &    0.5272      \\
                    & soft-label KD   & 0.4279     &   0.6249   & 0.3268      & 0.5278     \\
                    & $\text{\ours}^{r}$    &         0.4285      &      0.6398      & \textbf{0.3298}      &      0.5411      \\
                    & $\text{\ours}^{d}$    & \textbf{0.4301}      & \textbf{0.6496}     &         0.3271      & \textbf{0.5420}     \\ \midrule \midrule
                   & GPT-4                  & 0.3196               & 0.4939     & 0.3081               & 0.4395     \\ 
\bottomrule
\end{tabular}
}
\caption{EM and acc@5 scores of \ours on A-Cinderella and comparisons.}
\label{tab:main_aphaisa}
\vspace{-4mm}
\end{table}

The results are shown in Table~\ref{tab:main_aphaisa}. 
First, \ours improves performance over the $\text{DPR}_{maxsim}$. The performance follows the same trend with Subsec.~\ref{subsec:known-item-IR} ($\text{DPR}_{maxsim}$ $<$ Random $<$ $\text{\ours}_{C}$ $<$ hard- and soft-label KD $<$ \ours), across both noising methods and both datasets. This demonstrates that our incremental application over the original strategies provides significant improvements.

Second, our method holds \jongho{better} results compared to other circumlocution augmentation methods.
The performance comparison reveals that EDA which uses both random deletion and replacement performs worse than ours.
Moreover, neither of the approaches targeting only diversity, such as SMART and large-loss, nor those targeting only relevance such as small-loss has enhanced the performance from the random baseline. On the other hand, our strategy outperforms the baseline. This highlights leveraging both relevance and diversity is important in addressing anomia, which our strategy targets through the gradient-based selection. \ours outperforms VDA which also deals with both diversity and relevance.
We attached the significance test results in Appx.~\ref{appndx:significance}.

Finally, our model outperforms GPT-4 across both datasets. LLMs' shortcomings become evident when confronted with the tail data, and suffer from perturbations~\cite{Qiang2024PromptPC}. In contrast, our model effectively mitigates such shortcomings, underscoring its superior performance and robustness in handling data with unseen and \spe terms.


\subsection{Analysis}
\label{sec:discussion}
\label{subsec:gradientasaproxy}

\paragraph{Gradient as a Proxy for Perturbance}
We validate the alignment between the gradient order we have derived and the degree of SPE for each term. 

We assess whether terms with high gradients do not have SPE and are indeed relevant so that their removal will significantly degrade performance. 
To investigate this, we compare the effects of removing terms with different gradient degrees.
We rank the terms in the circumlocutions according to their gradient values, sorting them in descending order. We partition them into five intervals. We eliminate half of the terms within each interval and proceed with model training.

\begin{figure}[t]
    \centering
    \includegraphics[width=1.0\linewidth]{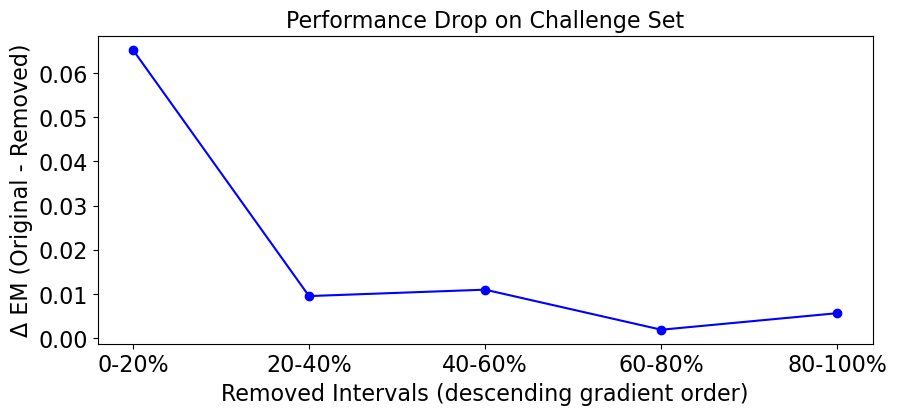}
    \caption{Quantitative analysis of the gradient-based proxy. The x-axis is the removed intervals based on gradient values and the y-axis is the EM score decrease. 
    }
    \label{fig:quantgradient}
\end{figure}

The result is shown in Fig.~\ref{fig:quantgradient}. It illustrates a notable decrease in model performance when terms with top gradients are removed. This confirms our hypothesis that top gradients serve as proxies for the not-perturbed ones, necessary for the model to identify the same target item with the noisy input consistently.

\paragraph{Relevance and Diversity of Gradient-based Selection}

\begin{table}[h]
\centering
\scalebox{0.8}{
\begin{tabular}{l|ccc}
\toprule
\textbf{m/n}  & Error rate (↓) & Distance (↑)  & Acc@5 (↑) \\
\midrule
0\%/0\% (Cutoff) & 0.3438 &  0.1476 & 0.5201 \\
5\%/0\% & \textbf{0.2714}  & 0.1446 & 0.5205 \\
0\%/70\% & 0.3630  & \textbf{0.1548}  & \underline{0.5252} \\
5\%/70\% (Ours) & \underline{0.2862}  & \underline{0.1540} & \textbf{0.5275}  \\
\bottomrule
\end{tabular}
}
\caption{Error rate, distance, and acc@5 results of ours compared to ablated on A-cinderella challenge set. The second-best score is underlined.}
\label{tab:rel_div}
\end{table}
Following~\citet{zhao-etal-2022-epida}, we verify the quality of augmented data from the perspective of relevance and diversity.  For relevance, we measure the augmentation error rate. It measures the percentage of augmented data that gets a lower acc@5 score than the original data. For diversity, we calculate the average cosine distance of the data before and after the augmentation. 
Additionally, we evaluated our ablated version ($\text{\ours}^{d}_{C}$) from the best hyperparameters and reported the performance. The results are demonstrated in Table~\ref{tab:rel_div}. 
Setting value $m > 0$ promotes relevance by preventing the keywords from noise injection, and value $n > 0$ promotes the diverse sample by targeting terms that affect the model performance. Ours improves the performance of the model by balancing diversity and relevance.
\section{Conclusion}

We studied designing LM to aid anomic patients by identifying the target item that they intend to.
We identified two primary challenges in this context raised from unseen and SPE terms. To tackle these issues, we proposed a gradient-based selective augmentation strategy, which robustifies the model from SPE terms and enhances it with unseen terms. Experiments show that addressing each challenge contributes to improving performance. We demonstrate that the gradient works as a proxy for SPE, showing its effectiveness in controlling the quality of data augmentation. \jongho{Our approach demonstrated consistent improvements in both anomic patients and healthy individuals experiencing Tip-of-the-Tongue states, thereby broadening the impact of this research not only on anomia but also on general studies of intended target identification.}
\section{Limitations}

Our study has several limitations that need to be addressed in future research.
Firstly, our approach relied on fine-tuning, which requires time and depends on hyperparameter optimization. 

\jongho{Second, while anomia presents a new challenge with many possible avenues for exploration, our focus was limited to leveraging the retrieval model with data augmentation. One unexplored direction is prompting LLMs, which we avoided due to GPT-4's low performance. Another approach could involve deleting SPE terms during inference. However, it is too challenging to delete them without knowing the target item. Thus, we concentrated on identifying non-perturbed terms during training and used an adversarial approach. These directions could offer valuable insights for future work.}

Lastly, while our experiments demonstrate the efficacy of our methods, the availability of retrieval failure speech from anomic patients remains a critical need. Access to such data would allow for a more comprehensive evaluation of our approach in real-world clinical settings.


\section*{Acknowledgement}

This work was supported by Institute of Information \& communications Technology Planning \& Evaluation (IITP) grant funded by the Korea government(MSIT) [NO.RS-2021-II211343, Artificial Intelligence Graduate School Program (Seoul National University)], and Institute of Information \& communications Technology Planning \& Evaluation (IITP) grant funded by the Korea government (MSIT) (No. 2022-0-00077/RS-2022-II220077, AI Technology Development for Commonsense Extraction, Reasoning, and Inference from Heterogeneous Data). We would like to thank Mr. Dohyeon Lee, our lab member, for his valuable discussions during the initial writing of this manuscript.

\bibliography{anthology,custom}
\bibliographystyle{acl_natbib}

\appendix

\section*{\centering Appendices}

\jongho{
\section{Related Works}

\subsection{Computational Approaches for Assessing PWA}
Aphasia encompasses various language impairments including anomia.
With the release of AphasiaBank~\cite{forbes2012aphasiabank}, several works have aimed to assist clinicians support PWA by detecting aphasia~\cite{le2017automatic, gale-etal-2022-post}, predicting aphasia severity~\cite{day2021predicting, wagner2023careful}, and the intended targets of aphasic speech. This paper concentrates on the last aspect.

While \citet{purohit2023chatgpt} tested the feasibility of ChatGPT in predicting these targets, such an investigation was constrained by a limited sample size that involved only 12 cases. Recently, \citet{salem2023automating} designed \jongho{a subcollection of AphasiaBank to evaluate the performance of LMs on aphasia.}
However, their evaluation setting requires prior knowledge of the answer length, which is impractical. 
We revised the dataset to target anomia and developed the LM that helps anomic patients in real-world scenarios.

\subsection{Information Retrieval}
In information retrieval, there are two primary approaches: traditional term-based methods like BM25~\cite{robertson2009probabilistic}, and modern vector-based methods like dense retrieval. 
\textbf{BM25} scores documents based on term frequency(how often a term appears in a document), inverse document frequency (how common a term is across the corpus), and document length normalization, making it a simple yet effective baseline for many retrieval tasks. Unlike BM25's reliance on exact term matching, dense retrieval methods, like \textbf{DPR} (Dense Passage Retrieval)~\cite{Karpukhin2020DensePR}, utilize dense vector representations to capture semantic meaning, typically generated by deep learning models such as BERT~\cite{Devlin2019BERTPO}. By encoding queries and documents into high-dimensional vectors and measuring their similarity, dense retrieval can outperform traditional methods like BM25, especially when queries and documents use different terms to express similar concepts. In concern of unseen terms, we leverage dense retrieval to identify anomia patients' intended target item.

\subsection{Data Augmentation}

Data augmentation strategies aim to introduce variations to the original data to improve the model's performance. Rule-based approaches include random deletion~\cite{shen2020simple, chen2021hiddencut}, replacement~\cite{wang-yang-2015-thats}, or both~\cite{wei2019eda}. However, these methods often fail to control the quality of the augmented data. To address, some techniques target diversity of the augmented data~\cite{zhu2019freelb, jiang2020smart,Yi2021ReweightingAS}, semantic relevance of the data~\cite{han2018co}, or both of them~\cite{zhou-etal-2021-virtual,zhao-etal-2022-epida}. We devised an augmentation strategy under the innate presence of perturbations, to guarantee diversity and relevance based on the observation that term gradients can serve as a proxy for perturbation.

\subsection{Self-Knowledge Distillation}

Knowledge distillation (KD)~\cite{Hinton2015DistillingTK} transfers knowledge from a teacher to a student model.  The conventional approach of KD utilizes high-capacity teachers and compact students. Self-knowledge distillation, on the other hand, is a variant of knowledge distillation where a model is trained to improve itself without the need for a separate teacher model. A line of research has demonstrated that students with parameters identical to their teachers can outperform the teachers themselves~\cite{furlanello2018born}. The effectiveness of self-KD has been validated across various tasks, including generation~\cite{wu2019exploiting, he2019revisiting} and information retrieval~\cite{kim2022collective,jiang2023noisy}. Our method differs from the original self-KD by selectively distilling the labels for the annotated data, to generate the relevance feedback~\cite{croft2010search} to the model.
}

\section{Qualitative Analysis of Perturbed Terms}
\label{sec:case_study_perturbed}

We selected a query of the test set for which the model ranks the target item \textit{Hero (2002 film)} as 14th. Below is the given query, and the terms that we considered perturbed are subsequently dropped:

\begin{tcolorbox}
I’m looking for a movie where a \sout{samurai} meets the emperor of \sout{Japan} and the movie cuts away to things this \sout{samurai} has done. And each thing he’s done allows the \sout{samurai} to move closer to the emperor and it’s all a plan for this guy to get close enough to \sout{kill} the emperor.
\end{tcolorbox}

We considered these terms perturbed because the plot is about \textit{assassins}, and is set in \textit{China}. When the perturbing terms are dropped, the model ranks the target item as top-1.

\section{\jongho{Experimental settings}}
\label{appndx:eval_details}
\subsection{Dataset Statistics}

Reddit-TOMT~\cite{bhargav2022s} consists of 13,253 known-item queries with 14,863 documents. Their queries and corresponding gold items are sourced from Tip-of-my-Tongue~\footnote{\url{https://www.reddit.com/r/tipofmytongue/}} Reddit sub-community. We use the official data split of train (80\%), validation (10\%), and test set (10\%).

TREC-TOT 2023~\cite{arguello2023overview} consists of 450 queries and 231,618 documents, extracted from ~\url{https://irememberthismovie.com}. We use an official 1:1:1 split for train, validation, and test sets. Evaluation is done with the pytrec package~\footnote{\url{https://github.com/cvangysel/pytrec_eval}}.

A-cinderella~\cite{salem2023automating} dataset comprises 353 Cinderella story sessions from 254 participants, with 2.5k unintended words. The item corpus is the lemmatized dictionary of all words within the dataset with a size of approximately 2k. We used a 10-fold cross-validation setting and reported the average score following~\citet{salem2023automating}.

\subsection{Evaluation Details}
For hyperparameters, the training was performed with a learning rate of $2e-5$ and a batch size of $16$. 
We search for $m$ in \{0.05, 0.1\} and $n$ in \{0.3, 0.5, 0.7\} for the circumlocution augmentation. In detail, we follow the convention in the field of adversarial attack~\cite{wu2023prada} for the threshold of high-gradient terms. They selected the top 50 high-gradient tokens for the adversarial attack, which is roughly 10\% considering the 512-token maximum of their BERT model. Following such value, we searched the hyperparameter for the high-gradient terms near the value 10\% (\{0.05, 0.1\}). For the threshold of low-gradient terms, we manually inspect the terms sorted by gradient. We found the threshold varies from case to case and selected a wider range for hyperparameter search (\{0.3 0.5 0.7\}). The best hyperparameters are in Table~\ref{tab:best_hyperparam}.
$\alpha$ and $\beta$ follow Cutoff(\citet{shen2020simple}). The value of $k$ is set to 2 for the item augmentation.

For known-item retrieval, we used $6$ RTX 3090 GPUs for $20$ epochs. For word completion, we used $1$ RTX 3090 GPUs for $10$ epochs.
The pre-trained LM in DPR is Bert-base-uncased.

Regarding A-cinderella, formulating the word completion as a retrieval task with DPR brings two benefits over the original MLM-based approach~\cite{salem2023automating}. First, MLM requires prior knowledge of the token length of the intended word, which is impractical, while our approach can assign a single mask token for any word, thus building the model for a more general purpose. Second, given the limited vocabulary of PWA, we can preset the search domain. 

\begin{table*}[h]
\centering
\scalebox{1.0}{
\begin{tabular}{l|ccc}
\toprule
\textbf{Hyperparameters} & \textbf{Reddit-TOMT} & \textbf{TREC-TOT} & \textbf{A-Cinderella (Both set)} \\
\midrule
$\alpha$  &  0.05  & 0.05  & 0.05 \\
$\beta$   &  0.7  &  0.5 & 0.3 \\
\bottomrule
\end{tabular}
}
\caption{The best hyperparameter values for circumlocution augmentation ($\alpha$, $\beta$).}
\label{tab:best_hyperparam}
\end{table*}

\subsection{Baselines Details}
\label{appndx:baseline_details}
\textbf{Cutoff~\cite{shen2020simple}} is a data augmentation strategy that involves selectively removing parts of the input data, such as spans, tokens, or dimensions to create new training examples. In Eq.~\ref{eq:cutoffloss}, $\alpha$ and $\beta$ are both selected from \{0.1, 0.3, 1\}. We note that it is the baseline that our code is built upon\footnote{\url{https://github.com/dinghanshen/Cutoff}}.

\textbf{EDA}~\cite{wei2019eda} involves simple operations such as synonym replacement, random insertion, random swap, and random deletion. These operations are applied to the text data to generate augmented datasets. The number of augmented sentences generated per original sentence is selected from \{1, 2, 3\}, with 1 yielding the best performance.

\textbf{SMART}~\cite{jiang2020smart} involves adding virtual adversarial noise to the input data to create diverse training examples. The loss is formulated as the minimax problem, in which the model is optimized to minimize the loss while simultaneously being robust against the perturbations introduced by the adversarial noise.
 The adversarial noise, $\delta$, maximizes the Kullback-Leibler ($\mathbf{KL}$) divergence between the original probability distribution $P(C,I)$ and yields the perturbed probability distribution $P_{\delta}(C_{aug},I)$:
\begin{align*}
\delta &= 
\underset{\lVert \delta \rVert_\infty \leq \epsilon}{\text{argmax}} \ \mathbf{KL}\left[ P \parallel P_{\delta} \right]
\end{align*}
Then the virtual adversarial loss, denoted as $\mathcal{L}{adv}$, is computed as the KL divergence between $P$ and $P_{\delta}$.
The variance of the noise $\delta$ is set to $1 \times 10^{-5}$, perturbation step $1$, and the step size $1 \times 10^{-3}$ for initializing perturbation.

\textbf{VDA}~\cite{zhou-etal-2021-virtual} is a virtual data augmentation strategy that targets both semantic relevance and diversity.
For semantic relevance, a masked language model is employed to generate virtual examples that are semantically similar to the original data. For diversity, Gaussian noise is incorporated into the augmentation process.
The variance of the Gaussian noise is set to $10^{-2}$ in our baseline.

\begin{figure*}[h]
    \centering
    \includegraphics[width=1.0\linewidth]{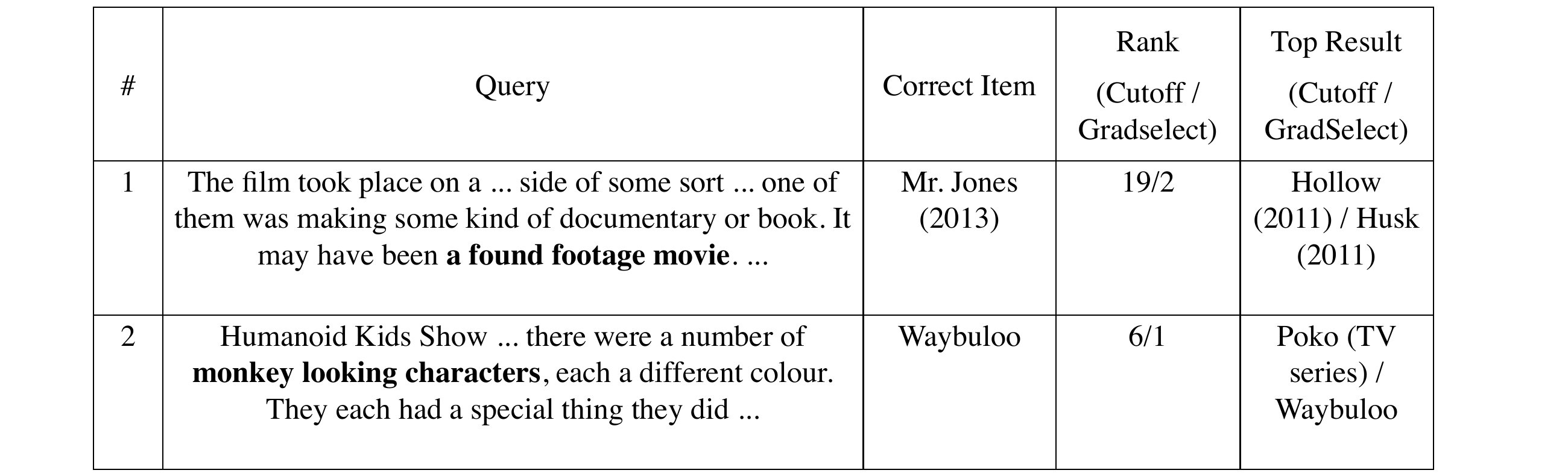}
    \caption{
    Case study results on Reddit-TOMT dataset. We compared the ranking results between Cutoff and GradSelect.
    }
    \label{fig:case_study_tomt}
\end{figure*}

\subsection{Prompts for GPT-4}
\label{appndx:gpt4_details}
\begin{table*}[h]
\centering
\begin{tabular}{l}
\toprule
\textbf{GPT-4 prompt Templates} \\
\midrule
System: You are helping the aphasia patient recall the Cinderella story. You respond to each message with \\
a list of 5 guesses for the word in [MASK]. **important**: you only mention the names of the words, \\
one per line, sorted by how likely they are the correct words with the most likely correct word first and \\
the least likely word last.
Do not output anything except for words. \\
User: \{$\cdot$\} \\
Assistant: \{$\cdot$\} \\
User:\{$\cdot$\} \\
Assistant: \{$\cdot$\} \\
User: \{$\cdot$\} \\
Assistant: \{$\cdot$\} \\
User: \{$\cdot$\} \\
Assistant: \\
\bottomrule
\end{tabular}
\caption{GPT-4 prompt templates used for A-Cinderella. \{$\cdot$\} is a placeholder for the random in-context examples.}
\label{tab:gptpromptcind}
\end{table*}
The GPT-4 prompt for A-cinderella is on Tab.~\ref{tab:gptpromptcind}. We evaluated GPT-4 in an in-context learning setting with a random 3-shot of examples. The prompt is based on GPT-4 prompt from TREC-TOT 2023~\cite{arguello2023overview}.

\subsection{Significance Test}
\label{appndx:significance}
We conducted paired t-tests on the challenge set of A-Cinderella using the acc@5 metric, with the test results across different folds in 10-fold cross-validation. GradSelect’s performance is significantly higher than the baseline $DPR_{maxsim}$, with $p<0.01$. For the significance test results of each component, our item augmentation is significantly better than both soft-label and hard-label KD, with $p<0.01$. Additionally, our circumlocution augmentation marginally outperforms the Cutoff (the best-performing baseline, $0.05<p<0.06$). The results are consistent for both $\text{GradSelect}^r$ and $\text{GradSelect}^d$.

\section{Case study}
We conducted case studies on the Reddit-TOMT test set, comparing the predictions from Cutoff and GradSelect. The Table~\ref{fig:case_study_tomt} shows the results. In the first query, there are several perturbed terms related to the film the main characters were making, such as `documentary/book/footage movie'. Cutoff wrongly retrieves the movie \textit{Hollow (2011)}, which itself is a found footage horror movie. In the second query, the user searches for `monkey-looking characters', and Cutoff retrieves the TV series \textit{Poko}, which features the `monkey cheracter'. In contrast, our approach correctly ranks the correct target item at the top.

\section{Usage of AI Assistants}
ChatGPT was employed to correct grammatical errors and to condense sentences to adhere to the page limit.

\end{document}